\documentclass[runningheads]{llncs}

\usepackage{graphicx}
\usepackage{booktabs}
\usepackage{tabularx}
\usepackage{array}
\usepackage{amsmath}
\usepackage{amssymb}
\usepackage[expansion=false]{microtype}
\usepackage[hidelinks]{hyperref}
\usepackage{tikz}

\usetikzlibrary{arrows.meta,positioning}
\makeatletter
\renewcommand{\paragraph}{\@startsection{paragraph}{4}{\z@}%
  {-6\p@\@plus-2\p@\@minus-1\p@}{-0.6em}{\normalfont\normalsize\itshape}}
\makeatother
\begin{document}

\title{Detecting and Repairing Hallucinations in
Retrieval-Augmented Generation}

\titlerunning{Detecting and Repairing Hallucinations in RAG}

\author{
Sai Krishna Reddy Mulakkayala\inst{1}
\and
Niki van Stein\inst{1}
\and
Aske Plaat\inst{1}
}

\authorrunning{S. K. R. Mulakkayala et al.}

\institute{
LIACS, Leiden University, Leiden, The Netherlands\\
\email{mulakkayalasaikrishnareddy@gmail.com\\
n.van.stein@liacs.leidenuniv.nl\\
aske.plaat@gmail.com}
}

\maketitle

\begin{abstract}
Language models increasingly answer questions by consulting retrieved documents
rather than memory alone, a design now common in search assistants and
enterprise knowledge tools. Grounding a model in retrieved text reduces
unsupported statements but does not eliminate them, and a reader cannot tell a
grounded sentence from an invented one. Most research on this problem stops at
detection, yet flagging a faulty answer changes nothing for the person reading
it, and little is known about which action should follow. Using RAGTruth, a
benchmark whose unsupported passages are annotated by hand, we split each
flagged answer into individual factual claims, check each against the retrieved
source, and compare leaving the answer untouched with three repair strategies of
increasing richness: deleting an unsupported claim, replacing it with source
text, and rewriting it. Three language models from different families judge the
916 repaired answers. Every strategy reduces the proportion of answers judged to
contain unsupported content, and all three judges agree on the ordering.
Deletion achieves the largest reduction while retaining least of the original
answer, at 64.3\% of the text, whereas rewriting retains 80.1\% and reduces
least. Repair is not confined to faulty answers: 83.5\% of answers annotated
clean are edited too. The strategies occupy different points on a
grounding preservation trade-off rather than forming a quality ranking, and
choosing between them needs evidence about answer usefulness that automatic
metrics cannot supply.

\end{abstract}

\section{Introduction}

When a language model answers a question about a document, the answer is
usually fluent, on topic and mostly correct. The difficulty is the remainder. A
date, a figure or a causal link may appear in the answer without appearing in
the document, and nothing in the writing marks it out. Retrieval-augmented
generation (RAG) was introduced to reduce exactly this~\cite{lewisrag}: a
retrieval component supplies relevant passages and the model writes while
looking at them. Grounding lowers the rate of invented content without removing
it~\cite{RAGTruth}. As these systems move into search assistants and enterprise
tools, that residual rate matters, because the reader cannot audit which
sentence came from the source.

Research on this failure has concentrated on finding the unsupported
statements. That is necessary but not sufficient: a flagged answer is still a
faulty answer until something acts on the flag. What the system should then do
is a separate question, and the options are not equivalent. Deleting the
offending statement is safe and discards information. Replacing it with source
text stays close to the evidence but can read awkwardly. Asking the model to
rewrite it keeps most of the answer and creates a fresh opportunity to be
wrong. Little evidence exists on how these choices compare, and this paper
supplies some.

The repair process studied here operates one claim at a time, where a claim is
a single factual statement extracted from the answer, so that a correction can
target the unsupported part of a sentence rather than the whole sentence. We
work on RAGTruth~\cite{RAGTruth}, which provides span-level hallucination
annotations across question answering, summarisation and data-to-text
generation.

We evaluate detector stability\footnote{A single-seed analysis, such as the seed-42 configuration is insufficient as a model-scaling result.
The repeated-seed experiment reported here supersedes single-seed experiments.} and a claim-level repair pipeline. Flagged answers are decomposed
into atomic claims, each claim is checked against the retrieved source, and
unsupported claims are either removed, replaced by source text, or rewritten.
We measure judged grounding together with claim and text retention.

We ask three research questions.

\begin{description}\setlength{\itemsep}{1pt}\setlength{\parskip}{0pt}

\item[\textbf{RQ1}]
Across repeated same-family adapter training, does the 13B detector reliably
outperform the 7B detector?

\item[\textbf{RQ2}]
How do removal-only, extractive and generative repair differ in judged
hallucination and content retention?

\item[\textbf{RQ3}]
How far do the repair conclusions depend on the automatic judge that measures
them?

\end{description}

\smallskip
\noindent\textbf{Contributions.}

\begin{enumerate}\setlength{\itemsep}{1pt}\setlength{\parskip}{0pt}

\item A five-seed QLoRA detector comparison showing that a favourable
single-run result does not establish a reliable model-size effect.

\item A controlled four-condition comparison of untouched answers and three
nested repair action sets: removal-only, extractive and generative.

\item A validation of the claim verifier against RAGTruth's human span
annotations, quantifying its ranking quality and its precision at the
operating threshold.

\item Evidence of a grounding preservation trade-off across three judge
families, together with a judge-reliability analysis that bounds how precisely
the automatic results can be interpreted.

\end{enumerate}

Detector scores are computed against RAGTruth's human annotations. In
contrast, repaired-answer hallucination labels come from prompted language
models. We therefore treat the automatic judges as measurement instruments
rather than as ground truth. No validated human or automatic answer-quality
evaluation is reported for the four-condition experiment.

\section{Related Work}

\paragraph{Hallucination and detection in RAG.}

Hallucination in grounded generation has been characterised and
benchmarked~\cite{jisurvey,RAGTruth}, and faithfulness studied in abstractive
summarisation~\cite{maynez}. Detection ranges from label-free consistency
checks~\cite{selfcheckgpt} to trained detectors. RAGTruth provides word-level
annotations over realistic RAG output, making both response-level detection and
span localisation measurable.

\paragraph{Recent detectors on RAGTruth.}

LettuceDetect trains a token-level hallucination detector on
RAGTruth~\cite{lettucedetect}. MiniCheck addresses the related task of checking
whether generated claims are supported by a grounding document~\cite{minicheck}.
We do not propose a new detector architecture or claim a direct
state-of-the-art comparison because these systems use different training data,
evaluation settings and metrics.

\paragraph{Verification and self-correction.}

A line of work prompts models to expose intermediate reasoning~\cite{cot},
verify and revise their answers~\cite{cove}, or critique evidence during
generation~\cite{selfrag}. These motivate our explicit verifier and repair
stages. We ask a complementary question: how does enlarging the repair action
set change grounding and retention?

\paragraph{Parameter-efficient fine-tuning and LLM-based evaluation.}

QLoRA and related adapter methods train few parameters on a quantised frozen
base~\cite{lora,qlora}, making detector scaling feasible on one GPU. Using large
language models as judges is now common~\cite{geval}, but their labels are not
human labels, so we use three judge families and quantify their agreement.

\section{Method}

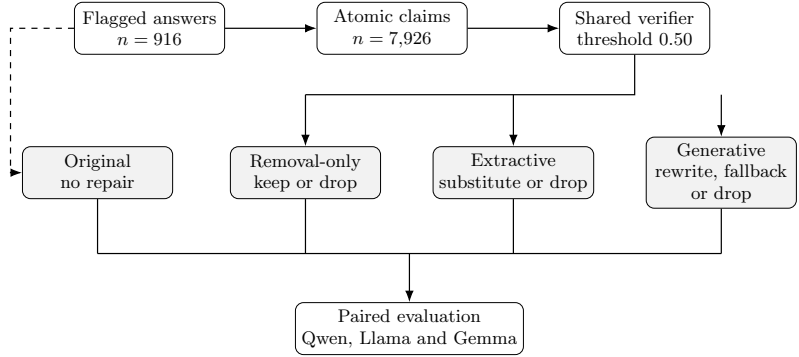
\begin{figure}[t]
\centering
\resizebox{0.86\linewidth}{!}{%
\begin{tikzpicture}[
  every node/.style={font=\footnotesize},
  box/.style={draw,rounded corners,align=center,minimum width=26mm,
    minimum height=9mm,fill=white,inner sep=3pt},
  arm/.style={box,fill=black!5},
  arr/.style={-{Latex[length=2mm]},semithick}
]
\node[box] (input)  at (0,0)    {Flagged answers\\$n=916$};
\node[box] (claims) at (4.2,0)  {Atomic claims\\$n=7{,}926$};
\node[box] (verify) at (8.4,0)  {Shared verifier\\threshold $0.50$};
\draw[arr] (input)  -- (claims);
\draw[arr] (claims) -- (verify);

\coordinate (bus) at (4.2,-1.15);
\draw[semithick] (verify.south) -- (8.4,-1.15) -- (2.7,-1.15);

\node[arm] (original) at (-0.9,-2.5) {Original\\no repair};
\node[arm] (remove)   at (2.7,-2.5)  {Removal-only\\keep or drop};
\node[arm] (extract)  at (6.3,-2.5)  {Extractive\\substitute or drop};
\node[arm] (generate) at (9.9,-2.5)  {Generative\\rewrite, fallback\\or drop};
\draw[arr] (2.7,-1.15)  -- (remove.north);
\draw[arr] (6.3,-1.15)  -- (extract.north);
\draw[arr] (9.9,-1.15)  -- (generate.north);

\draw[arr,dashed] (input.west) -- ++(-1.1,0) |- (original.west);

\coordinate (bus2) at (4.5,-3.9);
\draw[semithick] (original.south) -- (-0.9,-3.9);
\draw[semithick] (remove.south)   -- (2.7,-3.9);
\draw[semithick] (extract.south)  -- (6.3,-3.9);
\draw[semithick] (generate.south) -- (9.9,-3.9);
\draw[semithick] (-0.9,-3.9) -- (9.9,-3.9);
\node[box] (judges) at (4.5,-5.2)
  {Paired evaluation\\Qwen, Llama and Gemma};
\draw[arr] (bus2) -- (judges.north);
\end{tikzpicture}
}

\caption{
Four-condition paired design. Every answer appears unchanged and under each of
three nested repair action sets. Solid arrows carry claims routed by the shared
verifier; the dashed arrow indicates that the control condition receives the
answer unchanged and bypasses verification. All repair arms share the same
claim decomposition, eligibility scores, threshold and candidate-verification
caches, and all four conditions are evaluated by the same three judges.
}
\label{fig:pipeline}

\end{figure}

Figure~\ref{fig:pipeline} shows the evaluated pipeline.

\paragraph{Detector.}

The detector reads a source passage and generated response and produces a
response-level label together with predicted hallucination spans. The
detectors are Llama-2 models~\cite{llama2} trained with QLoRA. QLoRA updates
adapter layers on a quantised frozen base, allowing a 13-billion-parameter
detector to be trained on a single GPU.

\paragraph{Claim decomposition and verification.}

A flagged answer is split into atomic claims by a deterministic rule-based
splitter. Each claim is matched to source evidence through TF--IDF retrieval.
A prompted Qwen-2.5-32B verifier~\cite{qwen} scores each claim against the full
source context as $P(\mathrm{support}) = P(\text{``Yes''}) / (P(\text{``Yes''})
+ P(\text{``No''}))$, a renormalised two-token probability rather than a
calibrated posterior.

\paragraph{Repair actions.}

A claim is kept when $P(\mathrm{support}) \geq 0.50$ and enters repair
otherwise. The pipeline uses one routing threshold.

The three repair arms are nested. Removal-only drops every claim below the
threshold. The extractive arm first tries to substitute the highest-overlap
source fragment and drops the claim if that candidate is rejected. The
generative arm asks Qwen-2.5-7B to rewrite the claim; a rejected rewrite enters
the same extractive fallback and is dropped if that also fails. Before any
candidate is accepted, a do-no-harm gate checks for numbers, dates or named
entities absent from the source and verifies the edited text again.

\paragraph{A worked example.}

The following case is taken verbatim from the recorded outputs for record 29,
claim \texttt{r-s1-c0}, which the verifier scored at $P(\mathrm{support})=0.00$.

\smallskip
\noindent\begin{tabularx}{\linewidth}{@{}l>{\raggedright\arraybackslash}X@{}}
\textit{Source} & \dots She could be sentenced to 15 years in prison.\\
\textit{Claim} & She was arrested on March 26 and could face 15 years in
prison.\\
\textit{Removal-only} & (claim deleted)\\
\textit{Extractive} & She could be sentenced to 15 years in prison.\\
\textit{Generative} & She could face 15 years in prison.\\
\end{tabularx}

\smallskip
\noindent
The arrest date is unsupported and the sentence length is supported. Deletion
removes both; the two richer actions retain the supported half.

\paragraph{Automatic judges.}

Residual hallucination is scored using a prompted Qwen-2.5-32B judge that
labels a complete answer against its source. Because this model belongs to the
same family as the verifier, we add two judges from different model families:
Llama-3.1-70B and Gemma-2-27b-it.

Each judge emits hallucination spans for the complete answer. An answer is
labelled hallucinatory when at least one span is emitted. Retention measures
how much material survives repair. It does not indicate whether the retained
material is complete, helpful or fluent, and completeness, helpfulness and
fluency were not measured for this experiment.

\paragraph{Implementation validation.}

Before the final evaluation we validate the implementation against the
specification: that all model handles load, that every generation path produces
output on a smoke test, that the routing threshold is the one specified, that
record and claim keys are unique, and that saved actions reconstruct the emitted
responses. Hash, schema or model failures abort execution rather than becoming
repair decisions. Locking prompts and analyses is useful only once the relevant
code paths have been tested, because a component that fails to load leaves the
same trace as one with nothing to do.

\section{Experimental Setup}

\paragraph{Data and splits.}

All experiments use RAGTruth~\cite{RAGTruth}. The detector is trained on the
training split and evaluated on the official test split of 2{,}700 responses.
A 942-record validation split supports the span-level diagnostic.

The repair experiment uses a frozen intervention population of 916 test
responses containing 7{,}926 atomic claims. These are exactly the responses
that the seed-42 Llama-2-13B detector of Section~\ref{sec:detector} flagged as
containing unsupported content: the detector emits a boolean verdict per
response rather than a tunable score, and it flagged 916 of the 2{,}700 test
responses, a flag rate of 33.9\%. The two experiments therefore compose into a
single end-to-end system, with the caveat noted in Section~\ref{sec:detector}
that the selecting run is the best of five seeds.

Measured against the human annotations, that selection has precision 80.8\%
and recall 78.5\%. RAGTruth labels 740 of the flagged responses as
hallucinatory and 176 as clean; these are the true and false positives of the
selection step. A further 203 annotated-hallucinatory responses were not
flagged and are therefore never repaired, and 1{,}581 annotated-clean responses
were correctly left alone. The task mix of the flagged population is 578
data-to-text, 204 summarisation and 134 question-answering responses, which is
not the composition of the test split. The remaining 1{,}784 test responses are
outside the intervention analysis and pass through every repair condition
unchanged, so the results below describe the repair stage conditional on this
selection rather than end-to-end performance on the benchmark.

\paragraph{Evaluation protocol.}

The repair population, threshold, prompts, arm policies, outcomes and five
primary arm contrasts were fixed before outcome computation. Every record
appears in all four conditions. McNemar's exact test~\cite{mcnemar} is applied
to paired binary labels; paired bootstrap intervals use 10{,}000 record-level
resamples with seed 42; and $p$-values are Holm-corrected across the five
pre-specified contrasts within each judge. The additional removal-only versus
generative comparison is exploratory.

\paragraph{Models and training.}

The detector comparison holds the model family fixed, using Llama-2 at 7B and
13B under a 2{,}048-token context for three epochs. Both configurations share
the same family, context length, epoch count, data split and QLoRA protocol.

Each detector size is trained with seeds 13, 21, 42, 84 and 100. Matching seed
numbers do not create comparable initialisations across architectures, and the
realised cross-model correlation is negative. We therefore use Welch's
unpaired $t$-test rather than a paired analysis.

\section{Detection Results}
\label{sec:detector}

Response-level F1 measures whether a detector correctly identifies answers
containing at least one hallucination; span-level F1 measures how accurately it
locates the unsupported words. Table~\ref{tab:detector-seeds} reports every
response-level result from five training seeds per model size.

\begin{table}[t]
\centering
\caption{Response-level F1 for five training seeds per model size. Two
configurations failed at inference and are marked with a dagger: the 7B
seed-100 adapter produced empty generations on all 20 inspected examples with
no parser failures, and the 13B seed-13 failure was not diagnosed. Summary
statistics are given over all five runs and over the four usable runs only.}
\label{tab:detector-seeds}
\footnotesize
\begin{tabular}{lrrrrrrrrrr}
\toprule
& \multicolumn{5}{c}{Seed} & \multicolumn{2}{c}{All runs}
& \multicolumn{2}{c}{Usable runs} & \\
\cmidrule(lr){2-6}\cmidrule(lr){7-8}\cmidrule(lr){9-10}
Model & 13 & 21 & 42 & 84 & 100 & Mean & SD & Mean & SD & Median\\
\midrule
Llama-2-7B  & 78.5 & 78.8 & 66.0 & 76.0 & $1.3^{\dagger}$
            & 60.12 & 33.29 & 74.83 & 6.02 & 76.00\\
Llama-2-13B & $41.4^{\dagger}$ & 77.5 & 79.6 & 78.6 & 79.3
            & 71.28 & 16.72 & 78.75 & 0.93 & 78.60\\
\bottomrule
\end{tabular}
\end{table}

Two separate conclusions follow, and they should not be merged.

First, one run in five at each model size failed to produce usable generations
under this recipe. A targeted diagnostic of the 7B seed-100 adapter captured
empty generations on all 20 inspected examples, with no parser failures; the
training loss was comparable to successful runs, but the available artefacts
cannot distinguish an optimisation, serialisation, loading or generation
failure. The 13B seed-13 run was not diagnosed. The adapter checkpoints for both degraded runs were not retained after evaluation, so adapter weight norms cannot be inspected and the failure cannot be localised further. Because these runs produce no
usable output rather than poor output, the all-runs means of 60.12 and 71.28
mix a capability measurement with an inference-failure rate and are reported
for completeness only. Neither should be read as an estimate of what either
detector can do.

Second, among the four usable runs at each size the difference is small and not
statistically reliable. The 13B detector averages 78.75 F1 against 74.83 for
the 7B detector, a gap of 3.93 points with a 95\% confidence interval of
$-5.52$ to $13.37$ that includes zero, and Welch's unpaired test gives
$p=0.284$. Including the failed runs widens rather than resolves this: the
all-runs difference of 11.16 points carries a confidence interval of $-29.78$
to $52.10$ and $p=0.528$. The 79.6 score is therefore reported as the seed-42
result, not as evidence that the larger model reliably performs better.

The seed-42 run has a second role in this paper: its predictions define the
population repaired in Section~\ref{sec:repair}. We report this openly because
it cuts against the section's own conclusion. The repair experiment inherits a
selection made by the best of five runs, and a selection made by a median run
would have produced a different, and on this evidence probably noisier,
population.

\section{Verifier Validation}
\label{sec:verifier-validation}

The repair pipeline depends entirely on the verifier's claim-level decisions,
so we validate those decisions against RAGTruth's human span annotations. A
claim is labelled gold-unsupported when its character span overlaps any
annotated hallucination span in the same response, and the positive class is
``the verifier flags the claim as unsupported''.

\begin{table}[t]
\centering
\caption{Verifier performance against RAGTruth gold spans, over the 6{,}921
claims of 7{,}926 that could be aligned to character offsets. The operating
threshold used throughout the paper is 0.50.}
\label{tab:verifier-validation}
\footnotesize
\begin{tabular}{lrrrrrrr}
\toprule
Threshold & Flagged & TP & FP & FN & Precision & Recall & F1\\
\midrule
0.10 & 2{,}050 & 952 & 1{,}098 & 254 & 0.464 & 0.789 & 0.585\\
0.30 & 2{,}390 & 1{,}012 & 1{,}378 & 194 & 0.423 & 0.839 & 0.563\\
0.40 & 2{,}497 & 1{,}025 & 1{,}472 & 181 & 0.410 & 0.850 & 0.554\\
0.50 & 2{,}584 & 1{,}035 & 1{,}549 & 171 & 0.401 & 0.858 & 0.546\\
0.60 & 2{,}701 & 1{,}045 & 1{,}656 & 161 & 0.387 & 0.867 & 0.535\\
0.70 & 2{,}859 & 1{,}059 & 1{,}800 & 147 & 0.370 & 0.878 & 0.521\\
0.90 & 3{,}315 & 1{,}098 & 2{,}217 & 108 & 0.331 & 0.910 & 0.486\\
\bottomrule
\end{tabular}
\end{table}

Table~\ref{tab:verifier-validation} reports the result. We evaluated the
verifier against the human span annotations on the 6{,}921
claims of 7{,}926 that could be aligned to character offsets, of which 1{,}206
overlap an annotated hallucination span. Ranking quality is good: the area
under the ROC curve for $1-P(\mathrm{support})$ against the gold label is
0.881. Threshold behaviour is less favourable. At the operating threshold of
0.50 the verifier flags 2{,}584 claims, of which 1{,}035 are gold-unsupported
and 1{,}549 are not, giving precision 0.401 at recall 0.858, and F1 is highest
at the lowest threshold examined rather than at the operating point. The
alignment is approximate, because claims are produced by a rule-based splitter
while gold spans are annotated on the original response text, and 1{,}005
claims could not be aligned and are excluded; the reported precision therefore
describes only the alignable subset. The 2{,}584 claims flagged here are the
subset of the 3{,}014 repair-eligible claims that could be aligned; the
remaining 430 eligible claims fall outside the validated subset and are still
repaired.

\section{Repair Results}
\label{sec:repair}

\subsection{Execution and Repair Actions}

Table~\ref{tab:repair-actions} summarises the actions taken. The shared
verifier marks 3{,}014 of 7{,}926 claims as eligible for repair.
The removal-only arm drops all of them. The extractive arm accepts 766 source
substitutions and retains 5{,}678 claims in total. In the generative arm, the
rewriter is invoked for all 3{,}014 eligible claims, proposes 1{,}686 rewrites
and emits 913 rewrites that pass the final gate. Its fallback accepts 459
extractive substitutions, leaving 6{,}284 retained claims.

No generation is malformed, no model-error sentinel occurs, and no
infrastructure exception is converted into a repair decision. Both verifier
passes finish before arm assembly; assembly performs 5{,}004 read-only cache
lookups with zero misses, writes and model calls. All 916 records are emitted
under every condition.

\begin{table}[t]
\centering
\caption{Repair actions and retained content on the 916-answer intervention
population. Text retention is the mean record-level character ratio; brackets
give 95\% paired-bootstrap intervals.}
\label{tab:repair-actions}
\footnotesize
\begin{tabularx}{\linewidth}{>{\raggedright\arraybackslash}Xrrr}
\toprule
Quantity & Removal-only & Extractive & Generative\\
\midrule
Claims retained & 4{,}912 (62.0\%) & 5{,}678 (71.6\%) & 6{,}284 (79.3\%)\\
Text retained & 64.3\% [63.0,65.7] & 74.8\% [73.4,76.2] & 80.1\% [79.0,81.3]\\
Extractive substitutions & 0 & 766 & 459\\
Rewrites proposed / accepted & 0 / 0 & 0 / 0 & 1{,}686 / 913\\
Claims dropped & 3{,}014 & 2{,}248 & 1{,}642\\
\bottomrule
\end{tabularx}
\end{table}

\subsection{Worked Examples}
\label{sec:examples}

Table~\ref{tab:examples} shows one unsupported claim per task type, drawn
verbatim from the recorded outputs, with what each arm produced for it.

\begin{table}[t]
\centering
\caption{One unsupported claim per task type and the output of each arm, taken
verbatim from the recorded outputs. $P$ is the verifier's support probability
for the original claim.}
\label{tab:examples}
\footnotesize
\begin{tabularx}{\linewidth}{@{}l>{\raggedright\arraybackslash}X@{}}
\toprule
\multicolumn{2}{@{}l}{\textit{Summarisation, record 29, $P=0.00$}}\\
Source & \dots She could be sentenced to 15 years in prison.\\
Claim & She was arrested on March 26 and could face 15 years in prison.\\
Removal-only & (deleted)\\
Extractive & She could be sentenced to 15 years in prison.\\
Generative & She could face 15 years in prison.\\
\midrule
\multicolumn{2}{@{}l}{\textit{Data-to-text, record 5692, $P=0.01$}}\\
Source & \dots the menu is too scattered and overwhelming.\\
Claim & The ambiance is classy and casual, with upscale and trendy elements.\\
Removal-only & (deleted)\\
Extractive & Additionally, the menu is too scattered and overwhelming.\\
Generative & The ambiance is classy and casual.\\
\midrule
\multicolumn{2}{@{}l}{\textit{Question answering, record 12260, $P=0.03$}}\\
Source & \dots Step 4: scroll down the page to see Web browser section.\\
Claim & Scroll down to the Default browser section.\\
Removal-only & (deleted)\\
Extractive & Step 4: Now, on the right-side, scroll down the page to see Web
browser section.\\
Generative & Scroll down to see the Web browser section.\\
\bottomrule
\end{tabularx}
\end{table}

The cases illustrate the trade-off and one failure mode. In the first, the claim
mixes an unsupported date with a supported sentence length: deletion removes
both, the richer actions keep the supported half. In the third, the generative
arm corrects a wrong section name while keeping the instruction readable. The
second is a failure: the extractive arm substitutes a source sentence about the
menu for a claim about the ambiance, retaining characters while destroying the
meaning. Every metric in this paper counts that as retained content, which is
one reason retention must not be read as quality.

\subsection{Judged Hallucination}

\begin{figure}[t]
\centering
\includegraphics[width=0.68\linewidth]{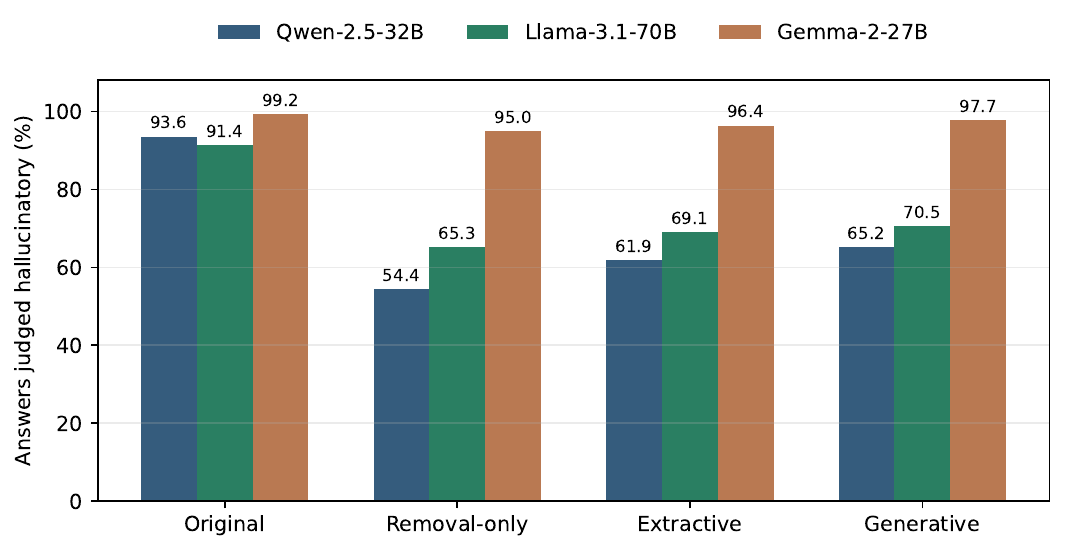}
\caption{Automatically judged hallucination rates. Every judge orders the
conditions identically, while their absolute estimates differ substantially.}
\label{fig:judge-rates}
\end{figure}

All three repair strategies lower the proportion of answers judged
hallucinatory relative to the untouched control under all three judges
(Table~\ref{tab:judge-rates} and Fig.~\ref{fig:judge-rates}). The absolute
levels differ sharply across judges: for example, the generative condition is
labelled hallucinatory for 65.2\% of answers by Qwen but 97.7\% by Gemma.

\begin{table}[t]
\centering
\caption{Answers labelled hallucinatory in each condition. Each cell reports
the percentage and count out of 916 answers.}
\label{tab:judge-rates}
\footnotesize
\begin{tabular}{lrrrr}
\toprule
Judge & Original & Removal-only & Extractive & Generative\\
\midrule
Qwen-2.5-32B  & 93.6 (857) & 54.4 (498) & 61.9 (567) & 65.2 (597)\\
Llama-3.1-70B & 91.4 (837) & 65.3 (598) & 69.1 (633) & 70.5 (646)\\
Gemma-2-27B   & 99.2 (909) & 95.0 (870) & 96.4 (883) & 97.7 (895)\\
\bottomrule
\end{tabular}
\end{table}

Relative to the original, the generative arm reduces judged hallucination by
28.38 percentage points under Qwen (95\% CI $[-31.55,-25.22]$,
Holm-corrected $p=7.0\times10^{-62}$), 20.85 points under Llama
($[-23.80,-17.90]$, $p=2.0\times10^{-41}$), and 1.53 points under Gemma
($[-2.51,-0.55]$, $p=0.013$). The corresponding reductions for removal-only
are 39.19, 26.09 and 4.26 points.

The pre-specified nested contrasts show that richer action sets leave more
judged hallucination. Extractive repair exceeds removal-only by 7.53 points
under Qwen, 3.82 under Llama and 1.42 under Gemma; all three Holm-corrected
tests are significant. Generative repair exceeds extractive repair by 3.27
points under Qwen and 1.31 under Gemma, while the 1.42-point Llama difference
is not significant ($p=0.130$). An exploratory direct comparison also finds
more judged hallucination under generative than removal-only for every judge.

\subsection{Behaviour by Task Type}
\label{sec:pertask}

The intervention population is 63.1\% data-to-text, so the aggregate figures are
dominated by that task. Table~\ref{tab:pertask} separates the three task types.

\begin{table}[t]
\centering
\caption{Judged hallucination and text retention by task type under the primary
Qwen judge. Eligible is the proportion of that task's claims scoring below the
fixed 0.50 threshold. Paired-bootstrap intervals on text retention are given in
the released result files.}
\label{tab:pertask}
\footnotesize
\begin{tabular}{lrrrrr}
\toprule
& \multicolumn{4}{c}{Judged hallucinatory} & \\
\cmidrule(lr){2-5}
Task & Orig. & Removal & Extract. & Gener. & Eligible\\
\midrule
Data-to-text ($n=578$)   & 96.7\% & 62.8\% & 69.7\% & 74.2\% & 36.6\%\\
Summarisation ($n=204$)  & 88.7\% & 40.7\% & 45.6\% & 47.5\% & 34.3\%\\
Question ans.\ ($n=134$) & 87.3\% & 38.8\% & 53.0\% & 53.0\% & 49.6\%\\
\midrule
& \multicolumn{4}{c}{Text retained} & \\
\cmidrule(lr){2-5}
Task & Orig. & Removal & Extract. & Gener. & \\
\midrule
Data-to-text   & 100\% & 65.5\% & 70.3\% & 76.9\% & \\
Summarisation  & 100\% & 67.6\% & 86.1\% & 89.5\% & \\
Question ans.\ & 100\% & 53.9\% & 77.0\% & 79.6\% & \\
\bottomrule
\end{tabular}
\end{table}

The ordering reported in Section~\ref{sec:repair} holds within every task, but
its size does not. On summarisation and question answering the primary judge
records reductions of 48.0 and 48.5 percentage points under removal-only; on
data-to-text it records 33.9. Data-to-text is also the task on which repair
retains least and on which the judges remain closest to their ceiling, and it
supplies 63.1\% of the population, so the aggregate figures are closer to the
data-to-text column than to the other two.

The single fixed threshold does not behave alike across tasks. It marks 49.6\%
of question-answering claims eligible for repair against 34.3\% of
summarisation claims, and the median support probability is 0.562 on question
answering against 0.971 on summarisation. The verifier is substantially less
confident on question answering, and the same threshold therefore routes a much
larger share of that task into repair. A per-task or calibrated threshold is a
natural next step, and is not evaluated here.

\subsection{Repair of Answers Annotated as Clean}
\label{sec:clean}

Repair is applied on the basis of detector and verifier decisions, not on the
basis of the benchmark's annotations, so it also edits answers that RAGTruth
records as containing no hallucination. Table~\ref{tab:clean-vs-hall} compares
the 176 such answers with the 740 annotated as hallucinatory.

\begin{table}[t]
\centering
\caption{Repair behaviour on answers RAGTruth annotates as clean ($n=176$) and
as hallucinatory ($n=740$). An answer counts as edited when at least one claim
is dropped or substituted. Claims removed is the mean per edited answer.
Brackets give 95\% paired-bootstrap intervals.}
\label{tab:clean-vs-hall}
\footnotesize
\begin{tabularx}{\linewidth}{>{\raggedright\arraybackslash}Xrrr}
\toprule
Quantity & Removal-only & Extractive & Generative\\
\midrule
\multicolumn{4}{l}{\textit{Annotated clean} ($n=176$)}\\
Answers edited & 83.5\% & 83.5\% & 83.5\%\\
Claims removed per edited answer & 2.69 & 1.75 & 1.28\\
Text retained & 74.6\% [71.4,77.6] & 87.2\% [84.4,89.7] & 90.0\% [87.6,92.3]\\
\midrule
\multicolumn{4}{l}{\textit{Annotated hallucinatory} ($n=740$)}\\
Answers edited & 98.1\% & 98.1\% & 98.1\%\\
Claims removed per edited answer & 3.61 & 2.74 & 2.00\\
Text retained & 61.9\% [60.4,63.4] & 71.9\% [70.3,73.4] & 77.7\% [76.4,79.0]\\
\bottomrule
\end{tabularx}
\end{table}

The pipeline edits 83.5\% of the answers annotated clean against 98.1\% of those
annotated hallucinatory, removing fewer claims per edited answer in the clean
subset (2.69 against 3.61 under removal-only) and retaining more text. The
judges nevertheless label most clean answers hallucinatory before any repair,
at 74.4\% for Qwen, 78.4\% for Llama and 97.7\% for Gemma, so reductions on this
subset partly reflect judge disagreement with the benchmark rather than the
correction of annotated errors. These data cannot establish whether the edits
remove unsupported content the annotation missed or damage sound answers.

\subsection{Grounding--Preservation Trade-off}

\begin{figure}[t]
\centering
\includegraphics[width=0.48\linewidth]{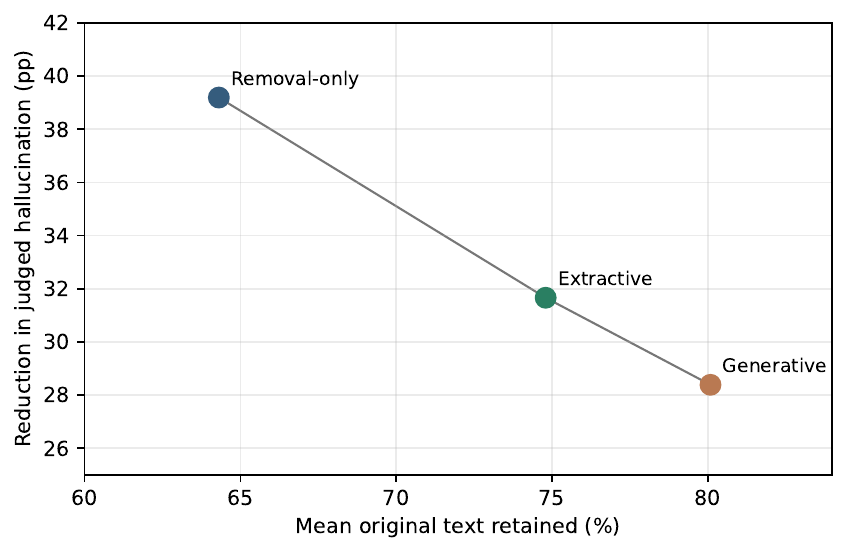}
\caption{Content retention versus reduction in judged hallucination under the
primary Qwen judge, with 95\% paired-bootstrap intervals on both axes. Moving
to richer repair actions preserves more text but reduces fewer hallucination
labels. Both axes are truncated to the observed data range in order to resolve
differences of a few percentage points; the intervals indicate the precision of
each point.}
\label{fig:tradeoff}
\end{figure}

The repair arms form a consistent trade-off. Removal-only produces the largest
reduction in judged hallucination but retains the least content. Generative
repair retains 80.1\% of the text and 79.3\% of claims, but produces the
smallest reduction. Figure~\ref{fig:tradeoff} shows this relationship under the
primary Qwen judge. The same ordering appears under the other judges.

This result does not identify which point on the trade-off is best for users.

\section{Judge Reliability}

All judges agree that every repair arm improves on the original and rank the
arms in the same order. However, their estimated effect magnitudes differ by
up to approximately a factor of nine. Pairwise Cohen's
$\kappa$~\cite{cohenkappa} ranges from 0.072 to 0.481 across conditions; no
pair reaches 0.60. On the original answers, agreement with RAGTruth's
response-level labels is also weak ($\kappa=0.317$ for Qwen, 0.203 for Llama
and 0.029 for Gemma). Specificity is the clearest symptom: of the answers the
benchmark annotates as clean, the judges correctly label only 25.6\%, 21.6\%
and 2.3\% respectively as clean.

The results therefore support the direction and ordering of the trade-off more
strongly than any single numerical estimate. In particular, Gemma labels more
than 94\% of every condition as hallucinatory, yielding much smaller absolute
changes than Qwen or Llama. Automatic judges are measurement instruments, not
substitutes for human evaluation.

\section{Discussion}

The repeated detector experiment changes the interpretation of the apparent
model-size advantage. Among usable runs the 13B mean exceeds the 7B mean by
3.93 F1 points, but the confidence interval includes zero and the test is not
significant. The two failed configurations show separately that a single
successful run can give an overconfident account of model scaling. Under this
training recipe, run-level variation is at least as important as the nominal
parameter count.

The repair comparison gives a clearer result. All three action sets improve
judged grounding relative to leaving the answers unchanged. However, no repair
strategy dominates both outcomes. Removal-only yields the lowest hallucination
rate, while generative repair retains the most claims and text. Extractive
repair lies between them. Generative rewriting should therefore not be
described as the best repair method: it represents a different operating point
on a grounding preservation trade-off.

No point on the trade-off can yet be recommended. A longer answer may preserve
useful detail or may preserve unsupported material and awkward edits, and
deletion can improve grounding while leaving an answer incomplete. The three
judges agree on the direction and ordering of the effects, which makes the
qualitative trade-off reasonably robust, but their low mutual agreement makes
any single percentage-point estimate hard to transfer. Human evaluation is
needed both to validate the labels and to establish which outcome readers
prefer.

The verifier validation locates the mechanism behind the pipeline's breadth of
intervention. The verifier ranks unsupported claims well, but at the fixed
operating threshold it flags a majority of claims that the human annotation
does not mark, and that low precision propagates directly into the editing of
answers annotated as clean reported in Section~\ref{sec:clean}.

\section{Limitations and Future Work}

All results are obtained on a single benchmark, and generalisation to other
domains, retrievers, generators and evidence distributions is not established.

The detector comparison contains only five runs per model size, of which one
per size failed at inference. Its confidence interval is correspondingly wide.
The available artefacts do not identify whether the failures arose during
optimisation, serialisation, loading or generation. The results therefore
characterise this particular QLoRA recipe, not QLoRA in general.

The repair evaluation covers detector-flagged answers only. The selection step
misses 203 answers that carry a human hallucination annotation, and those are
never repaired, so no end-to-end effect on the full test split is reported
here. The population is also enriched by construction, at 80.8\% annotated
hallucinatory against a 34.9\% base rate in the split, and skewed towards
data-to-text. Results on a differently selected population may differ.

The verifier operates at low precision at the fixed threshold (0.401 at recall
0.858), so a majority of claims entering repair are not gold-unsupported, and
the validation itself excludes the 1{,}005 claims that could not be aligned to
gold spans.

Text and claim retention measure how much material remains, not whether it is
complete, helpful, coherent or fluent, and no human or automatic quality
evaluation was completed. The study compares grounding against retention; it
cannot claim that any condition is better for readers.

Residual hallucination is measured entirely by automatic judges whose pairwise
agreement is low and whose effect sizes differ substantially, so their labels
are model-derived measurements rather than ground truth.

A blinded multi-annotator study should measure factual support, completeness,
helpfulness, fluency and overall preference for the same four outputs. Other
priorities are replication on a second benchmark, more detector runs with
checkpoint retention, and explicit optimisation of the grounding--preservation
trade-off.

\section{Conclusion}

We evaluated hallucination detection and three claim-level repair strategies
for retrieval-augmented generation. Across five training seeds per model size,
one run per size failed at inference, and among usable runs the 13B detector
exceeds the 7B detector by 3.93 F1 points without statistical reliability. The
experiment therefore does not establish a consistent benefit from increasing
detector size.

On 916 detector-flagged answers, removal-only, extractive and generative repair
all reduce automatically judged hallucination relative to the original under
three judge families. Removal-only produces the largest reduction, whereas
generative repair retains the most content. The methods thus occupy different
points on a grounding--preservation trade-off rather than forming a simple
quality ranking. Repair is not confined to faulty answers: 83.5\% of the
answers the benchmark annotates as clean are also edited, and the verifier's
precision of 0.401 against gold spans indicates why.

The judges agree on this ordering but disagree substantially on magnitude and
show low pairwise agreement. Together with the absence of answer-quality and
human evaluation, this prevents a claim about which strategy is best for
readers.

The wider implication concerns how systems of this kind are assessed. A repair
pipeline can be made to look arbitrarily good on a hallucination metric by
deleting more of the answer, and the metric will not object. Grounding and
usefulness are separate axes, and measuring only the first rewards a system for
saying less. Whether the content that rewriting preserves is worth its higher
residual hallucination rate is a question about readers, and answering it
requires blinded human annotation measuring grounding and usefulness together.
Until that evidence exists, systems of this kind should be reported as
operating points rather than as improvements.

\begin{credits}

\subsubsection{Acknowledgements}

The authors thank SURF for access to the Snellius national supercomputer and
Leiden University for access to the ALICE compute cluster.

\subsubsection{\discintname}

The authors have no competing interests to declare.

\end{credits}

\subsubsection*{Reproducibility.}

All reported quantities, the prompts, the scripts used to compute them and the
raw model outputs are available at
\url{https://github.com/Saikrishna-cy/rag-hallucination-repair}.
Each reported number is traceable to a named result file and the script that
produced it.

\appendix

\section{Reproducibility Details}
\label{app:repro}

Both detectors are Llama-2 base models (7B and 13B) fine-tuned with QLoRA on a
4-bit NF4 quantised frozen base, at LoRA rank 32, alpha 64, learning rate
$5\times10^{-4}$, three epochs, a 2{,}048-token context and an effective batch
size of 16. Seeds 13, 21, 42, 84 and 100 are used for detector training and 42
for bootstrap resampling. Checkpoints are selected on development data, and each
trained run receives one test evaluation.

The verifier is Qwen-2.5-32B, prompted and quantised to 4 bits, routing on a
single support threshold of 0.50 fixed in advance. The repair planner is
Qwen-2.5-7B, invoked on every eligible claim in the generative arm. The three
judges are Qwen-2.5-32B, Llama-3.1-70B and Gemma-2-27b-it, running the same
prompt with no per-judge tuning. Verifier and judge decoding is greedy at
temperature 0.

Detector training used one A100 40\,GB. Verifier, rewriter and judge inference
used one or two A100 GPUs with 40 to 80\,GB depending on model size; arm
assembly and all statistics ran on CPU. Exact prompts, per-record result files
and the scripts that recompute every reported number are in the supplementary
material.

\begingroup
\scriptsize
\let\small\scriptsize
\setlength{\itemsep}{-3pt}
\setlength{\parskip}{0pt}

\bibliographystyle{splncs04}
\bibliography{references}

\endgroup

\end{document}